\documentclass[journal]{IEEEtran}
\usepackage{times}
\usepackage{epsfig}
\usepackage{graphicx}
\usepackage{amsmath}
\usepackage{amssymb}
\usepackage{multirow}
\usepackage{booktabs}       
\usepackage{epstopdf}
\usepackage{verbatim}

\usepackage[pagebackref=true,breaklinks=true,letterpaper=true,colorlinks,bookmarks=false]{hyperref}
\begin{document}

\title{Diversity-Achieving Slow-DropBlock Network for Person Re-Identification}

\author{Xiaofu~Wu$^\dag$, Ben~Xie, Shiliang~Zhao, Suofei~Zhang, Yong~Xiao and Ming~Li

\thanks{$^\dag$Corresponding author. This work was supported in part by the National Natural Science Foundation of China under Grants 61372123, 61671253 and by the Scientific Research Foundation of Nanjing University of Posts and Telecommunications under Grant NY213002.}
\thanks{Xiaofu~Wu, Ben~Xie and Shiliang~Zhao are with the National Engineering Research Center of Communications and Networking, Nanjing University of Posts and Telecommunications, Nanjing 210003, China (E-mails: xfuwu@ieee.org, \{1018010631, 1018010632\}@njupt.edu.cn).}
\thanks{Suofei Zhang is with the School of Internet of Things, Nanjing University of Posts and Telecommunications, Nanjing 210003, China (E-mail: zhangsuofei@njupt.edu.cn).}
\thanks{Yong Xiao is with the School of Electronic Information and Communications, Huazhong Univ. of
Science \& Technology, Wuhan 430074, China (E-mail: yongxiao@hust.edu.cn).}
\thanks{Ming~Li is with the Supply Chain Platform Division, Alibaba Group, Hangzhou 311121, China
(Email: sebastian.lm@alibaba-inc.com).}}

\maketitle

\begin{abstract}
   A big challenge of person re-identification (Re-ID) using a multi-branch network architecture is to learn diverse features from the ID-labeled dataset. The 2-branch Batch DropBlock (BDB) network was recently proposed for achieving diversity between the global branch and the feature-dropping branch. In this paper, we propose to move the dropping operation from the intermediate feature layer towards the input (image dropping). Since it may drop a large portion of input images, this makes the training hard to converge. Hence, we propose a novel double-batch-split co-training approach for remedying this problem. In particular, we show that the feature diversity can be well achieved with the use of multiple dropping branches by setting individual dropping ratio for each branch. Empirical evidence demonstrates that the proposed method performs superior to BDB on popular person Re-ID datasets, including Market-1501, DukeMTMC-reID and CUHK03 and the use of more dropping branches can further boost the performance.
\end{abstract}


\newtheorem{plm}{Problem}
\newtheorem{thm}{Theorem}
\section{Introduction}

Person re-identification (Re-ID) focuses on matching images associated with the same person taken by the same or different cameras at different angles, time or location. It has attracted significant interest due to its fundamental role in emerging computer vision applications such as surveillance, human identity validation, and authentication, and human-robot interaction \cite{zheng2016person,dai2019BDB,chen2019MHN,yang2019CAMA,hou2019IAN,chen2019ABD,luo2019bag,TMM2019GLAD,TMM2020FFLN}. Despite its enormous progress, identifying the person of interest accurately and reliably is still very challenging under huge variations in lighting, human pose, background, camera viewpoint, etc. With an end-to-end training approach, one of the main goals in the field of person Re-ID is to discover useful features as rich as possible from the labeled dataset.

\begin{figure}[t]
\centering
\includegraphics[width=8.0cm]{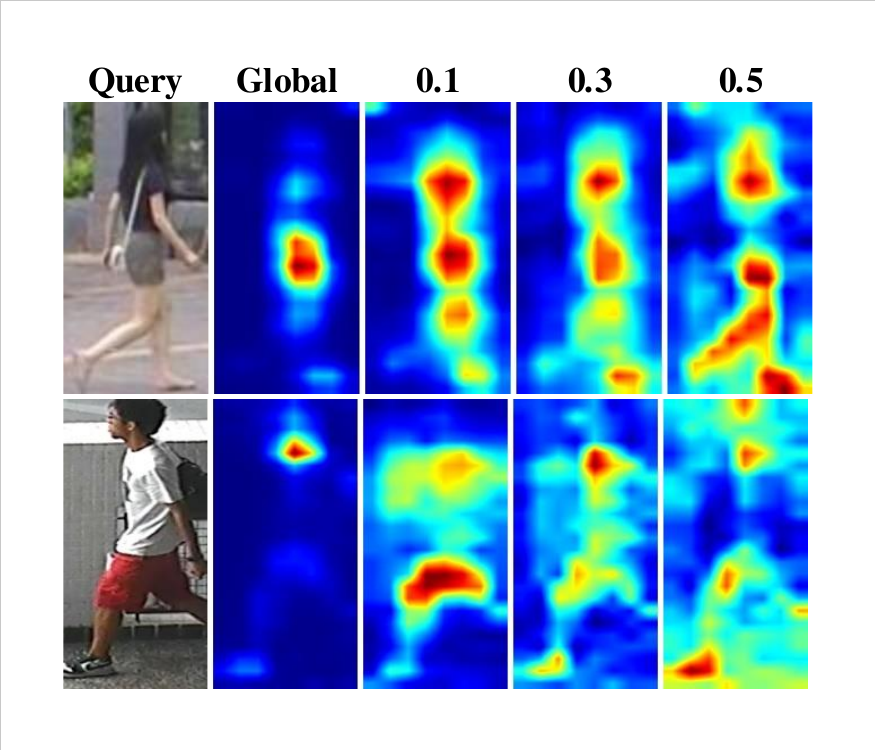}
\caption{Visualization of Class Activation Maps (CAMs) for diverse SDB branches. The proposed local dropping branches with different dropping ratios ($r_h=0.1, 0.3, 0.5$), along with the global branch for SDB allow the model to learn diverse features (marked in orange).}
\label{fig:dem}
\end{figure}

In general, person Re-ID can be considered as a feature-embedding (or metric-learning) problem~\cite{su2017pose,chen2017beyond,bai2017reid}, where the distance between
intra-class samples (associated with the same person) should be less than the distance between inter-class ones (associated with different persons) by at least a margin. Unfortunately, most existing feature-embedding solutions require grouping samples in a pairwise manner, which is known to be computationally intensive. In practice, the classification task is often employed to find the feature-embedding solution due to its clear advantage on the implementation complexity for training. Today, most state-of-the-art methods \cite{he2019Bag,dai2019BDB,chen2019ABD,zheng2017discriminatively,hermans2017defense} for person Re-ID have evolved from a single metric-learning problem or a single discriminative classification problem to a multi-task problem, where both the discriminative loss and the triplet loss are employed~\cite{su2016deep}. As each sample image is only labeled with the person ID, an end-to-end training approach usually has difficulty to learn diverse and rich features without elaborate design of the underlying neural network and further use of some regularization techniques.

In the past years, various part-based approaches~\cite{yao2017deep,sun2018beyond,zhao2017deeply} and dropout-based approaches~\cite{dai2018batch} have been proposed in order to learn rich features from the ID-labeled dataset. Differing from conventional pose based Re-ID approaches~\cite{su2017pose,kumar2017pose,zheng2017pose,qian2018pose}, part-based approaches usually locate a number of body parts firstly, and force each part meeting individual identification loss in order to get discriminative part feature representations \cite{wang2018MGN,suh2018part,cheng2016person,fan2018scpnet}. Dropout-based approaches, however, intend to discover rich features from enlarging the dataset with various dropout-based data-augmentation methods, such as cutout \cite{DeVries2017Cut} and random erasing \cite{zhong2017Erasing}, or from dropping the intermediate features from feature-extracting networks, such as Batch BropBlock \cite{dai2019BDB}.

\textbf{Drawbacks of Part-based Methods}: The performance of part-based methods relies heavily on the employed partition mechanism. Semantic partitions may offer stable cues to good alignment but are prone to noisy pose detections, as it requires that human body parts should be accurately identified and located.  The uniform horizontal partition is widely employed in \cite{sun2018beyond,suh2018part}, which, however, provides limited performance improvement, along with multi-branch network architecture because of the lack of semantic support and the difficulty in determining the appropriate number of partitions.

\textbf{Drawbacks of Dropout-based Methods}: Dropout-based methods have been widely used in person Re-ID, including various data dropping methods (for the purpose of data augmentation) and feature dropping methods. Data dropping methods, such as random erasing \cite{zhong2017Erasing}, cutout \cite{DeVries2017Cut} and DropBlock \cite{ghiasi2018dropblock}, have been shown to be effective for extracting rich features for person Re-ID. One of the main drawbacks is that the proportion of dropping pattern should be kept small enough for ensuring the convergence of training, which may hamper the discover of more diverse features. As a typical feature dropping method,  Batch DropBlock (BDB) \cite{dai2019BDB} has been proven to be effective for person Re-ID. However, the dropping pattern in BDB is fixed within only one iteration (one batch of samples), where the network may have difficulty in learning the corresponding structure. One way to improve the diversity of feature discovery is to increase the number of branches. Unfortunately, BDB is restricted to a two-branch architecture. Currently, it remains unknown on how to extend the existing two-branch architecture to the architecture with an arbitrary number of branches, e.g., $L$-branch ($L\ge 2$), for achieving improved diversity.

This motivates the work in this paper where we propose a novel multi-branch architecture for discovering rich features in person Re-ID. We briefly summarize the main contribution of this paper as follows:
\begin{enumerate}
\item
Based on 2-branch BDB network, we propose to move the dropping operation from the intermediate feature layer towards the input image. In particular, we propose a Slow-DropBlock (SDB) method, a slower version of Batch DropBlock, in constructing a multi-branch network for person Re-ID. Our method allows the dropping pattern keeping fixed over a number of ($Q$) batches, which facilitates the learning of stable features with extended time duration.

\item
By dropping a large portion of input images, the proposed SDB often makes the training process hard to converge. To address this issue, we propose a novel double-batch-split co-training approach with guaranteed convergence.

\item
\textit{One of main challenges in designing multi-branch networks for person Re-ID is to ensure feature diversity among individual branches. We show that feature diversity can be achieved with the use of multiple SDB branches by setting individual dropping ratio for each branch, as demonstrated in Figure \ref{fig:dem}}.

\item
The proposed SDB, along with double-batch-split co-training approach, has been proved very efficient for achieving the state-of-the-art performance on the three popular person Re-ID datasets, Marktet1501, DukeMTMC-reID and  CUHK03. For example, SDB achieves the rank-1 accuracy of 90.2\% for DukeMTMC-reID without using re-ranking \footnote{Source codes are available at \href{url}{https://github.com/AI-NERC-NUPT/SDB}.}

\end{enumerate}

\section{Related Work}
\label{gen_inst}
\subsection{Variation of Dropout Techniques}
Dropout is a standard tool for avoiding overfiting~\cite{srivastava2014dropout}, which randomly discards the output of each hidden neuron with a probability during training and forces the neural network to learn more diverse features. In recent years, novel dropout-based techniques have been proposed and further adopted in the field of person Re-ID.

\textbf{Cutout}: Cutout\cite{DeVries2017Cut} is a simple data augmentation technique that involves removing contiguous sections of input images, effectively augmenting the dataset with partially occluded versions of existing samples. It randomly masks out square regions of input during training.

\textbf{Random Erasing}: Instead of generating occluded samples, Random Erasing \cite{zhong2017Erasing} randomly selects a rectangle region in an image and erases its pixels with random values.

\textbf{DropBlock}: For a batch of input tensors (samples or features), DropBlock \cite{ghiasi2018dropblock} randomly drops a contiguous region for each input tensor.

\textbf{Spatial Dropout}: Spatial Dropout \cite{Tompson2015SpatialDropout} randomly zeroes whole
channels of the input tensor. The channels to zero-out are randomized on every forward call.

\textbf{Batch DropBlock}: Batch DropBlock \cite{dai2019BDB} randomly drops the same region of a batch of input feature maps and reinforces the attentive feature learning of the remaining parts.

Note that among various dropout methods, Spatial Dropout \cite{Tompson2015SpatialDropout}, and Batch DropBlock \cite{dai2019BDB} are performed over intermediate feature maps, while Cutout, Random Erasing, and DropBlock \cite{ghiasi2018dropblock}, are data-augmentation methods over input images.

\subsection{Two-Branch vs. Multi-Branch for Diverse Person Re-ID}
To get diverse features from an end-to-end training approach, multi-branch network architectures have been widely employed~\cite{sun2018beyond,dai2019BDB,chen2017multi}, where a shared-net is followed by  $L$ branches of subnetworks for achieving diversity in feature spaces. It is a natural problem to ask how many branches are necessary for person Re-ID. However, there is no easy answer to this question.

\textbf{Two-Branch Network ($L=2$)}: Two-branch network architectures, such as BDB \cite{dai2019BDB}, ABD \cite{chen2019ABD} and SONA\cite{xia2019SONA}, have been successfully employed for person Re-ID. All these networks consist of two branches: a global branch and a local branch. To achieve feature diversity between branches, distinct mechanisms should be imposed over the global and local branches, such as attention \cite{chen2019ABD} and feature dropping \cite{dai2019BDB,xia2019SONA}.

\textbf{Multi-Branch Network ($L\ge 2$)}: PCB \cite{sun2018beyond} employed a 6-branch network by dividing the whole body into 6 horizontal stripes. Following PCB, MGN \cite{wang2018MGN} employed a multi-branch network architecture consisting of one branch for global feature representations and two branches for local feature representations. A 4-branch network architecture was adopted in Auto Re-ID \cite{quan2019Auto}. MHN \cite{chen2019MHN} employed 6-branch high-order attention modules for achieving best possible performance. CAMA \cite{yang2019CAMA} used a 3-branch network by forcing an end-to-end overlapped activation penalty for achieving the best possible performance.  The number of branches is 21 in Pyramid \cite{zheng2019pyramid} based on a basic 6 horizonal stripes of PCB.

Despite of the recent efforts on the multi-branch network architecture, there is no clear advantage for adopting over the two-branch counterpart. For example, The 2-branch solution of SONA \cite{xia2019SONA} is superior to the 6-branch solution of MHN \cite{chen2019MHN}, both focusing on the attention mechanism. Hence, it remains to be exploited for how to obtain rich feature diversity for a multi-branch solution with $L\ge 2$.

\begin{figure}[t]
\begin{center}
\includegraphics[width=7.0cm]{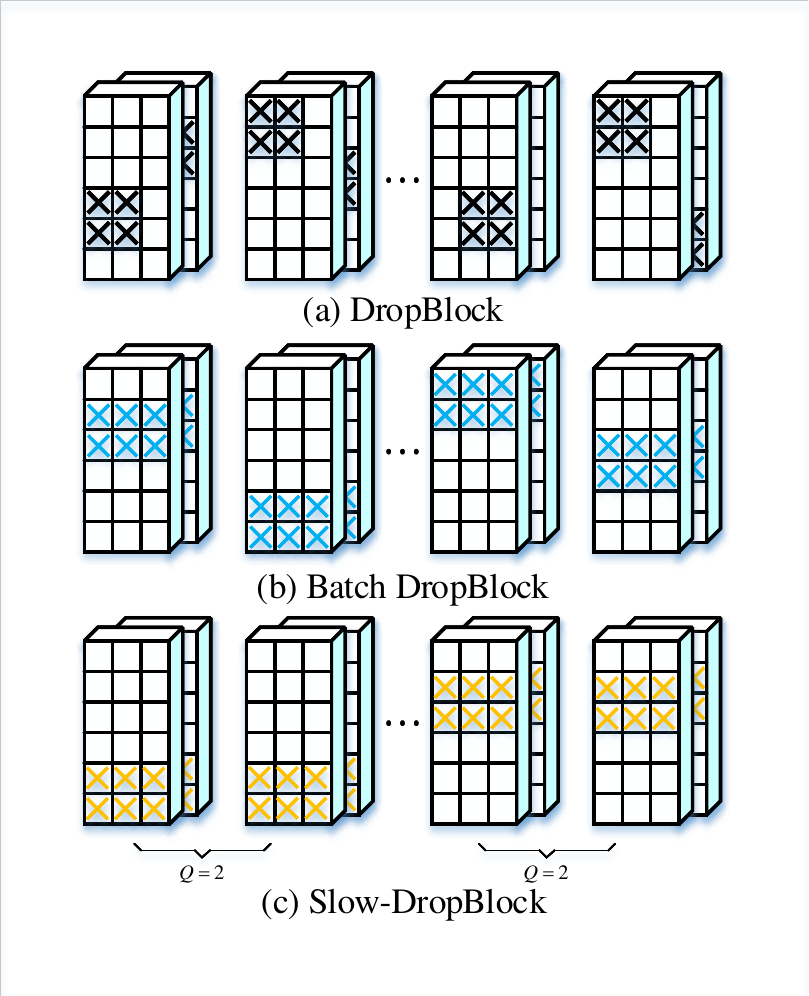} 
\end{center}
   \caption{Comparison of (a) DropBlock, (b) Batch DropBlock, and (c) Slow-DropBlock ($Q$=2) for a batch size of 2.}
\label{fig:DataAug}
\end{figure}

\begin{figure*}
\begin{center}
\includegraphics[width=0.95\textwidth]{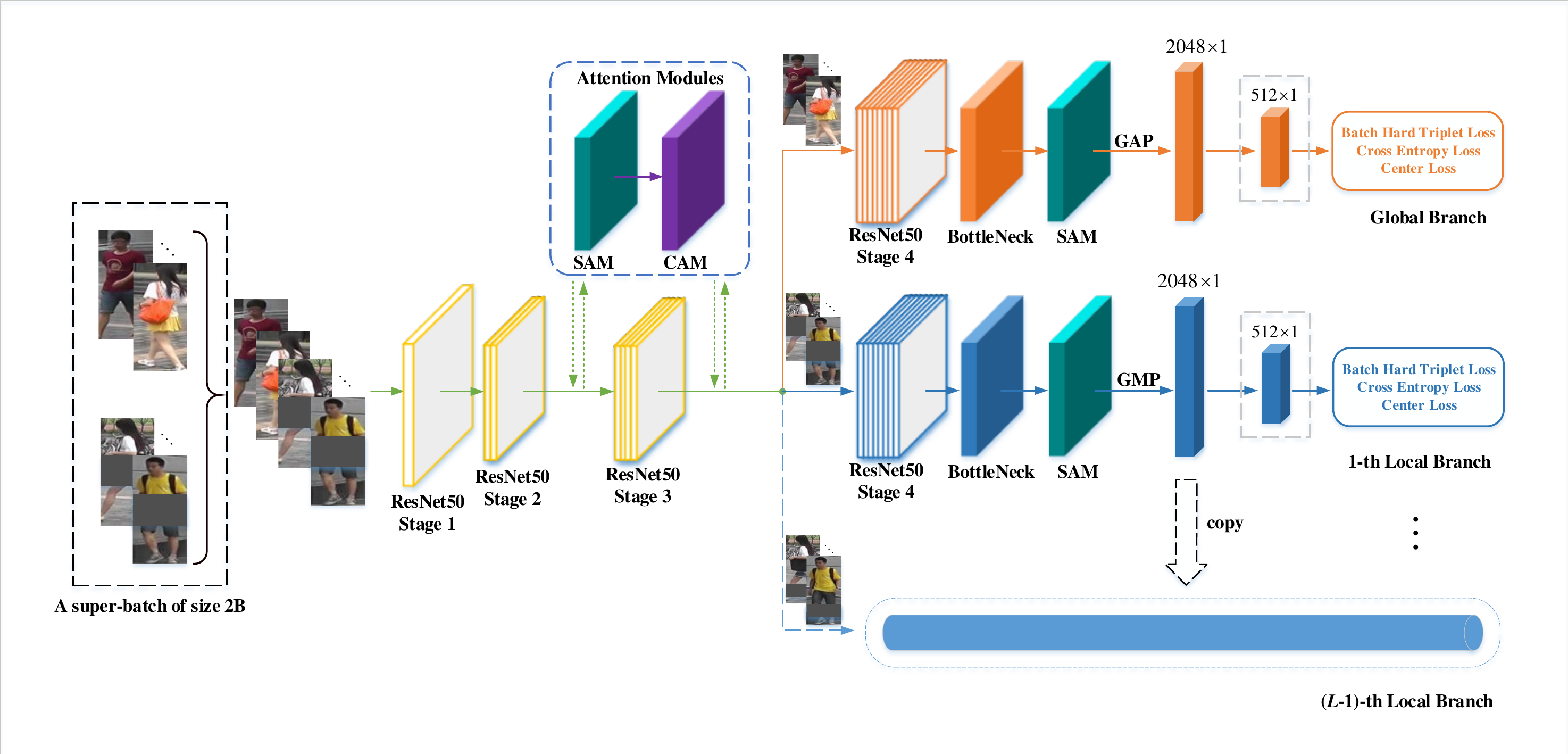} 
\end{center}
   \caption{The overall $L$-branch network architecture for the proposed SDB-$L$. With double-batch-split co-training for SDB-2, a super batch of $2B$ images are input to the network, among which half of them are conventionally augmented for global branch, and the other half of them are further augmented  with Slow-DropBlock for local branch. During testing, the feature embedding concatenated from both global branch and local branch is used for the final matching distance computation.}
\label{fig:net}
\end{figure*}

\section{Slow-DropBlock Network with Double-Batch-Split Co-Training}
\label{headings}

\subsection{Slow-DropBlock}
As shown in Figure \ref{fig:DataAug}, Slow-DropBlock randomly generates a contiguous block pattern and erases its pixels for $Q$ batches of input images.  ``slow'' means that the generated block patten keeps unchanged for at least $Q\ge 1$ batches. In the case of $Q=1$, this is equivalent to BDB \cite{dai2019BDB} for the purpose of data augmentation.

The only parameter in SDB is the erased height ratio $r_h$. In this way, the erased width ratio $r_w$ is always set to 1.

\subsection{Proposed Network Architecture}
As shown in Figure \ref{fig:net}, we employ an $L$-branch neural network architecture, called SDB-$L$, by modifying the commonly-used ResNet-50 baseline. Figure \ref{fig:net} shows the overall network architecture, which includes a backbone network, a global branch (orange colored arrows), and $L-1$ local branches (blue colored arrows). The main difference between SDB-2 and BDB is the use of attention modules in our proposed architecture. In this paper, the use of both spatial and channel attention modules in the shared-net follows the work of \cite{hou2019IAN} and we, however, insert an extra Spatial Attention Module (SAM) in the subsequent global and local branches. For the use of dimension reduction layer, experiments show that it depends on the underlying dataset.  It is required for CUHK03, but not necessarily for both Market-1501 and DukeMTMC-reID.

\subsubsection{Attention Modules}
Given an input feature map $X \in \mathbb{R}^{h\times w \times c}$, where $h$, $w$ and $c$ denote the height and the width of the feature map, the number of channels, respectively. Following the work in \cite{vaswani2017Attention, chen2019ABD,xu2018attention}, the employed spatial-attention module (SAM) first computes a spatial correlation matrix of $X$
 \begin{eqnarray}
 \Xi= X (I - \beta \mathbf{1}) X^T,
 \end{eqnarray}
where the input feature map is reshaped into a tensor $X$ of size $l\times c$ with $l=h\times w$ and $\beta$ is a parameter. Throughout this paper, we simply set $\beta =0$.  Then, the affinity matrix can be obtained as
\begin{eqnarray}
 \Lambda= \frac{\exp(\Xi)}{\|\exp(\Xi)\|_1}.
\end{eqnarray}
Then, the output of spatial attention module is
\begin{eqnarray}
 Y = X + \lambda \Lambda X,
\end{eqnarray}
where $\lambda$ is a learnable parameter.

Compared to SAM, the channel attention module (CAM) performs similarly, but focusing on the channel dimension instead of the spatial dimension.

\subsubsection{Shared-Net}
We use the popular ResNet-50 as the backbone network for feature extraction. For a fair comparison with the recent works \cite{dai2019BDB,chen2019MHN,yang2019CAMA}, we also modify the backbone ResNet-50 slightly, in which the down-sampling operation at the beginning of stage 4 has been removed. In this way, we get a larger feature map of size $2048 \times 24 \times 8$. As shown in Figure \ref{fig:net}, we insert SAM + CAM  modules in both stages of 3 and 4 for the shared-net.

\subsubsection{Global Branch}
The global branch consists of the stage-4 layers, a bottleneck block, a SAM module and a global average pooling (GAP) layer to produce a 2048 dimension vector and a feature reduction module containing a $1\times 1$ convolutional layer, a 2-D batch normalization layer, and a ReLU layer to reduce the dimension to 512 providing a compact global feature representation for both triplet loss and cross entropy loss.

\subsubsection{Diverse Local SDB Branches}
The local branch has the similar layer structure but with the global max pooling (GMP) in replace of GAP. Furthermore, the  reduction module follows the GMP layer is a $2048 \times 512$ fc-layer, followed by a batch-normalization layer and a ReLU layer.

To achieve feature diversity over local branches, we consider to employ multiple local branches with different settings of erased height ratio $r_h$. \textit{A main contribution of this paper is that we show that feature diversity can be achieved with multiple local SDB branches of different $r_h's$}. For example, with the setting of $r_h=0.2, 0.3, 0.4$ for 3 local SDB branches, we can concatenate features from 3 local branches with the global branch for achieving feature diversity.

\subsubsection{Loss Functions}
The feature vectors from the global and local branches are concatenated as the final embedded feature for the person
Re-ID task.  The loss function is the sum of identification loss (softmax loss), soft margin batch-hard triplet loss~\cite{hermans2017defense} and center loss~\cite{wen2016discriminative} on both the global branch and local branch, namely,
\begin{eqnarray}
 L_{total} = L_{id} + \gamma_t L_{triplet} + \gamma_c L_{center},
\end{eqnarray}
where $\gamma_t, \gamma_c$ are weighting factors.

\subsection{Double-Batch-Split Co-Training}
Firstly, we focus on the training of a $2$-branch SDB network (SDB-2). Experiments show that the use of SDB over input batches of images make the training process difficult to converge. Therefore, we propose a novel double-batch-split co-training scheme to address this problem.  During training, the network accepts a super-batch of images with size $2B \times H \times W \times 3$, where a super-batch is composed of 2 batches of samples, one usual batch and one dropping batch, both of size $B$. The usual batch is with the conventional data augmentation, including horizontal flip, normalization, and cutout. The dropping batch is further with the SDB data augmentation, where the dropping pattern is randomly generated but keeps fixed for $Q$ independent batches.

This super-batch is first input to the shared-net, including stage-1 to stage-4 layers. The resulting super-batch of features are then split into two sub-batches, one for the global branch, the other for the local branch. The obtained feature for each individual branch will then be of batch size $B$. The global branch uses the global average pooling (GAP)  while the global max pooling (GMP) is employed in the local branch, because GMP encourages the network to identify comparatively weak salient features after the most discriminative part is dropped.

Secondly, for an SDB-$L$ with $L\ge 2$, we run the above co-training process $L-1$ times, each for an individual SDB-2 nework, composed of the global branch and the $l$-th local branch, $l=1,\cdots, L-1$.

During testing, features from the global branch and all the local branches are concatenated as the embedding vector of a pedestrian image.

\section{Experiments}
Extenstive experiments have been performed for evaluating the effectiveness of the proposed approach over three public person Re-ID datasets: Market-1501, DukeMTMC-reID and CUHK03. The results are compared to the state-of-the-art methods.

\subsection{Datasets}

The Market-1501 dataset  \cite{zheng2015scalable} has 1,501 identities collected by six cameras and a total of 32,668 pedestrian images. Following \cite{zheng2015scalable}. The dataset is split into a training set with 12,936 images of 751 identities and a testing set of 3,368 query images and 15,913 gallery images of 750 identities.

The DukeMTMC-reID dataset \cite{Ristani2016Performance} contains 1,404 identities captured
by more than 2 cameras and a total of 36,411 images. The training subset contains 702 identities with 16,522 images and the testing subset has other 702 identities.

The CUHK03 dataset \cite{Li2014DeepReID} contains labeled 14,096 images and detected 14,097 images of a total of 1,467
identities captured by two camera views. With splitting just like in \cite{zheng2015scalable}, a non-overlapping 767 identities are for training and 700 identities for testing. The labeled dataset contains 7,368 training images, 5,328 gallery, and 1,400 query images for testing, while the detected dataset contains 7,365 images for training, 5,332 gallery, and 1,400 query images for testing.

\subsection{Implementation Details}
Our network is trained using a single Nvidia Tesla P100 GPU with a batch size of 32 ($B=32$). Hence, the super-batch is with size of $2B=64$. Each identity contains 4 instance images in a batch, so there are 8 identities per batch. The backbone ResNet-50 is initialized from the ImageNet pre-trained model. We use the batch hard soft margin triplet loss. The total number of epochs is set to 120 [150], namely, 120 for both Market-1501 and DukeMTMC-reID,  and 150 for CUHK03, respectively.  We use the Adam optimizer with the base learning rate initialized to 3.5e-5. With a linear warm-up strategy in first 10 [40] epochs, the learning rate increases to 3.5e-4. Then, the learning rate is decayed to 3.5e-5 after 40 [100] epochs, and further decayed to 3.5e-6 after 65 [125] epochs. For SDB, we use the default setting of $Q=5$. We use $r_h=0.3$ for SDB-2 and $r_h=0.2,0.3,0.4$ for SDB-4.

For training, the input images are re-sized to $384\times 128$ and then augmented by random horizontal flip, cutout, random erasing, and normalization. The testing images are re-sized to $384 \times 128$ with normalization.

\subsection{Comparison with State-of-the-art Methods}

We compare our work with state-of-the-art methods, in particular emphasis on the recent remarkable works (CVPR'19 and ICCV'19) on person Re-ID, over the popular benchmark datasets Market-1501, DukeMTMC-ReID and CUHK03. All reported results are obtained without any re-ranking \cite{zhong2017re,saquib2018pose} or multi-query fusion \cite{zheng2015scalable} techniques. The comparison results are listed in Table 1, Table 2 and Table 3. From these tables, one can observe that our proposed method performs competitively among various state-of-the-art methods, including Beyond Part models (PCB) \cite{sun2018beyond}, Batch BropBlock (BDB) \cite{dai2019BDB}, Mixed High-Order Attention Network (MHN) \cite{chen2019MHN}, CAMA \cite{yang2019CAMA}, Interaction-and-Aggregation Network (IAN)\cite{hou2019IAN}, ABD-Net \cite{chen2019ABD}.

As shown, SDB-4 performs consistently better than SDB-2, which means that feature diversity can be well achieved with the use of a multi-branch SDB architecture. For DukeMTMC-reID, our SDB-4 performs the best among various state-of-the-art methods, which has  1.2/3.5\% improvements of Rank-1/mAP over BDB, demonstrating the effectiveness of the proposed SDB.
By pushing DropBlock operation from intermediate features to inputs, SDB shows its superiority over BDB \cite{dai2019BDB} for all datasets. Compared to SONA, another variant of BDB with attention modules, SDB performs still better on all datasets.

\begin{table}
\caption{Comparison of our proposed method with state-of-the-art methods for the Market-1501 dataset}
\begin{center}
\label{my-label}
\begin{tabular}{l||cc}
\toprule[1.5pt]
Method     & mAP & rank-1  \\\hline
KPM \cite{shen2018KPM}(CVPR'18)       & 75.3 & 90.1   \\
MLFN \cite{chang2018MFN}(CVPR'18)     & 74.4 & 90.0   \\
CRF \cite{chen2018CRF}(CVPR'18)       & 81.6 & 93.5   \\
PCB \cite{sun2018beyond}(ECCV'18)       & 81.6 & 93.8  \\
SNL \cite{li2018SNL}(ACM'18) & 73.43 & 88.27  \\
HDLF\cite{zeng2018HDLF}(ACM MM'18) & 79.10 & 93.30  \\
MGN \cite{wang2018MGN}(ACM MM'18) & 86.9 & 95.7  \\
Local CNN\cite{yang2018LCNN}(ACM MM'18) & 87.4 & 95.9  \\
IAN \cite{hou2019IAN} (CVPR'19)       & 83.1 & 94.4  \\
CAMA \cite{yang2019CAMA}(CVPR'19)       & 84.5 & 94.7 \\
MHN \cite{chen2019MHN}(CVPR'19)       & 85.0 & 95.1   \\
Pyramid~\cite{zheng2019pyramid}(CVPR'19) & 88.2 & 95.7 \\
SONA \cite{xia2019SONA} (ICCV'19)       & 88.67 &95.68  \\
ABD \cite{chen2019ABD} (ICCV'19)       & 88.28 &95.6  \\
BDB \cite{dai2019BDB} (ICCV'19)       & 86.7 & 95.3  \\\hline
SDB-2       & 88.2 & \ 95.7  \\
SDB-4       & \bf88.7 & \bf95.9 \\
\bottomrule[1.5pt]
\end{tabular}
\end{center}
\label{tb:Market-1501}
\end{table}

\begin{table}
\caption{Comparison of our proposed method with state-of-the-art methods for the DukeMTMC-reID dataset}
\begin{center}
\label{my-label}
\begin{tabular}{l||ccc}
\toprule[1.5pt]
Method     & mAP & rank-1   \\\hline
MLFN \cite{chang2018MFN}(CVPR'18)     & 62.8 & 81.2   \\
GP-Re-ID \cite{almazan2018re} (CVPR'18)  & 72.8 & 85.2 \\
PCB \cite{sun2018beyond} (ECCV'18)       & 69.2 & 83.3  \\
MGN \cite{wang2018MGN}(ACM MM'18) & 78.40 & 88.7 \\
Local CNN\cite{yang2018LCNN}(ACM MM'18) & 66.04 & 82.23 \\
IAN \cite{hou2019IAN} (CVPR'19)       & 73.4 & 87.1  \\
CAMA \cite{yang2019CAMA} (CVPR'19)       & 72.9 & 85.8 \\
MHN \cite{chen2019MHN} (CVPR'19)       & 77.2 & 89.1  \\
SONA \cite{xia2019SONA} (ICCV'19)       & 78.05 &89.25  \\
BDB \cite{dai2019BDB} (ICCV'19)       & 76.0 & 89.0  \\\hline
SDB-2       & 78.9 & 89.8 \\
SDB-4       & \bf79.5 & \bf90.2 \\
\bottomrule[1.5pt]
\end{tabular}
\end{center}
\label{tb:Duke}
\end{table}

\begin{table}
\caption{Comparison of our proposed method with state-of-the-art methods for the CUHK03 dataset}
\begin{center}
\begin{tabular}{l|c@{\hskip 5pt}c@{\hskip 5pt}|c@{\hskip 5pt}c@{\hskip 5pt}}
\toprule[1.5pt]
\multirow{2}{*}{Method}	&	\multicolumn{2}{c|}{Labeled}	&			\multicolumn{2}{c}{Detected}			\\
\cline{2-5}
	&	mAP 	&	rank-1	&	mAP 	&	rank-1	\\
\hline
DaRe+RE~\cite{wang2018resource}(CVPR'18)	&	61.6 	&	66.1	&	59.0	&	63.3 \\
MLFN~\cite{chang2018MFN}(CVPR'18)	&	49.2 	&	54.7	&	47.8	&	52.8	\\
HA-CNN~\cite{li2018harmonious}(CVPR'18)	&	41.0 	&	44.4	&	38.6	&	41.7	\\
Local-CNN~\cite{yang2018LCNN}(ACM MM'18)	&	53.83 	&	58.69	&	51.55	&	56.76	\\
PCB~\cite{sun2018beyond}(ECCV'18) &    -   &   -   &	57.5	&	63.7	\\
MGN~\cite{wang2018MGN}(ACM MM'18)	&	67.4 	&	68.0 	&	66.0 	&	68.0 	\\
MHN \cite{chen2019MHN} (CVPR'19)   &72.4    & 77.2 & 65.4 & 71.7 \\
Pyramid~\cite{zheng2019pyramid}(CVPR'19)	&	76.9	&	78.9	&	74.8	&	78.9	\\
SONA \cite{xia2019SONA} (ICCV'19)       & 79.23 & 81.85 & 76.35 & 79.10   \\
BDB~\cite{dai2019BDB} (ICCV'19)	&	76.7	&	79.4	&	73.5	&	76.4	\\
\hline
SDB-2 	&	78.0	&	80.4	& 74.8	& 77.2	\\
SDB-4 	&	\textbf{80.7}	& \textbf{82.6}	& \textbf{77.4}	& \textbf{79.5}	\\
\bottomrule[1.5pt]
\end{tabular}
\end{center}
\label{tbl:CUHK03}
\end{table}

\subsection{Visualization}
\begin{figure}[t]
\centering
\includegraphics[width=8.0cm]{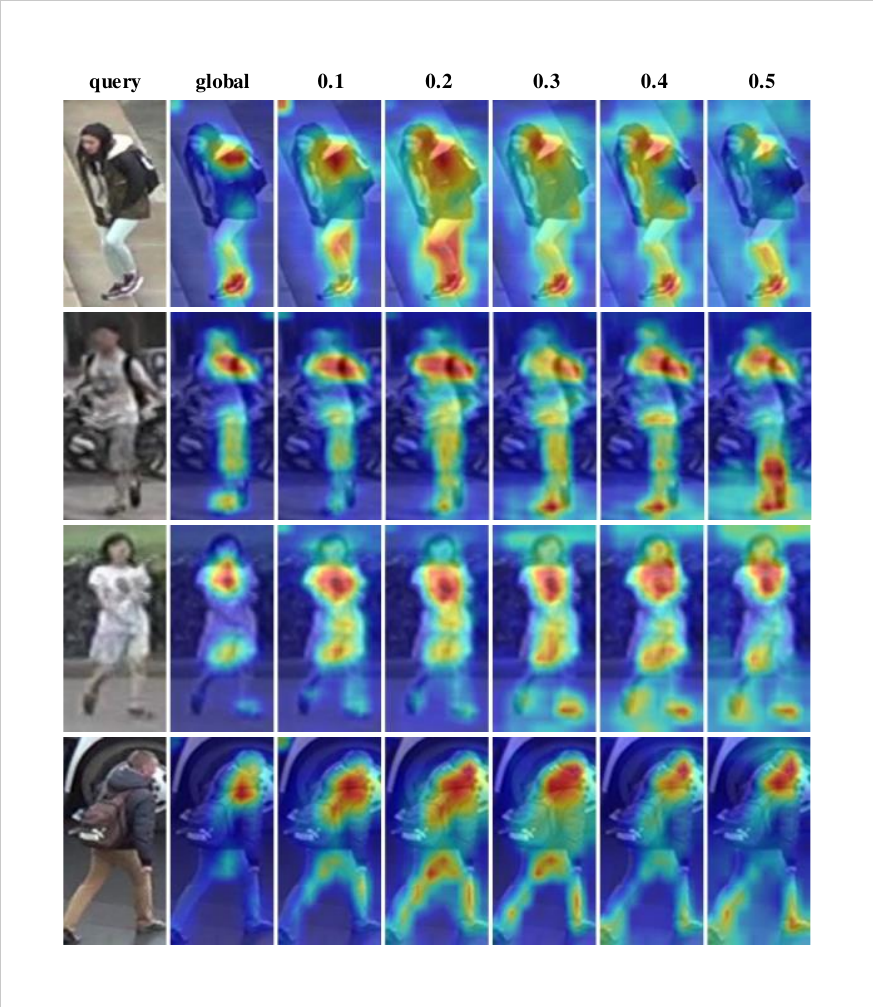}
\caption{Visualization of class activation maps (CAMs) for a global branch and multiple diverse local SDB branches. The proposed local branches with different dropping ratios ($r_h=0.1,0.2,0.3,0.4,0.5$) for SDB allow the model to learn diverse features (marked in orange).}
\label{fig:demDetail}
\end{figure}

\begin{figure}[t]
\centering
\includegraphics[width=8.0cm]{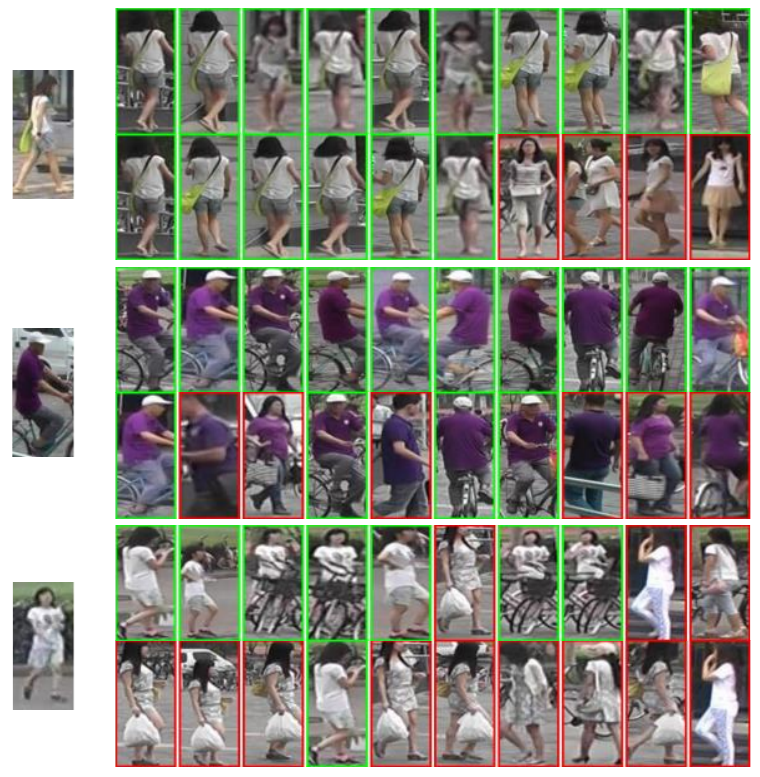}
\caption{Three Re-ID examples of SDB-2 and BDB on Market-1501. Left: query image. Upper-Right: top-10 results of SDB. Low-Right: top-10 results of BDB. Images in red boxes are negative results. SDB-2 boost the retrieval performance}
\label{fig:visualDem}
\end{figure}

\textbf{Diverse Feature Visualization}: We show that diverse features can be clearly observed with the use of the proposed local SDB branches with different dropping ratios in Figure \ref{fig:demDetail}, where 5 local branches of $r_h=0.1,0.2,0.3,0.4,0.5$ and the global branch are highlighted.

\textbf{Re-ID Visual Retrieving Results}:
We compare SDB-2 with BDB more directly from visual retrieving results. Three retrieved examples are shown in Figure \ref{fig:visualDem}. We can see that BDB fails to retrieve several correct images among the top-10 results.  Taking the first query as an example, SDB-2 is able to find correct images of the same identity in the top 10 results whilst BDB gets 4 incorrect ones.

\subsection{Ablation Studies}
\begin{table*}[ht]
\caption{Comparison with Batch BropBlock \cite{dai2019BDB} under the same SDB-2 network architecture}
\begin{center}
\begin{tabular}{l|c@{\hskip 5pt}c@{\hskip 5pt}|c@{\hskip 5pt}c@{\hskip 5pt}|c@{\hskip 5pt}c@{\hskip 5pt}|c@{\hskip 5pt}c@{\hskip 5pt}}
\toprule[1.5pt]
\multirow{2}{*}{Method} &	\multicolumn{2}{c|}{Market-1501}	 &	\multicolumn{2}{c|}{DukeMTMC} &	\multicolumn{2}{c|}{CUHK03-Labeled}	&			\multicolumn{2}{c}{CUHK03-Detected}			\\
\cline{2-9}
	&	mAP 	&	rank-1 &	mAP 	&	rank-1  &	mAP 	&	rank-1	&	mAP 	&	rank-1	\\
\hline\hline
BDB	&	83.9	&	94.2 &	76.5	&	88.6 &	76.8	&	79.2	& 72.3 & 76.0	\\
\hline
SDB-2 &	\textbf{88.2} &	\textbf{95.7} & \textbf{78.9} &	\textbf{89.8} &	\textbf{78.0}	&	\textbf{80.4}	& \textbf{74.8}	& \textbf{77.2}	\\
\bottomrule[1.5pt]
\end{tabular}
\end{center}
\label{tb:BDBcompare}
\end{table*}
\subsubsection{Comparison with BDB}
Since the proposed SDB is very similar to BDB, it is interesting to compare them in a fair way. Hence, we perform experiments over the SDB-2 network architecture by placing the Batch DropBlock before the GMP module on the local branch, just as did in BDB. Instead of using double-batch-split co-training, we employ the regular single-batch training approach in BDB. The results are shown in Table \ref{tb:BDBcompare}. As shown, SDB-2 clearly outperforms BDB for all the three datasets. For example, the improvement for SDB-2 over BDB in mAP is about 4.3\% for Market-1501.

\subsubsection{Benefit of Co-Training}
\begin{table*}[ht]
\caption{Benefit of co-training over three datasets}
\begin{center}
\begin{tabular}{l|c@{\hskip 5pt}c@{\hskip 5pt}|c@{\hskip 5pt}c@{\hskip 5pt}|c@{\hskip 5pt}c@{\hskip 5pt}|c@{\hskip 5pt}c@{\hskip 5pt}}
\toprule[1.5pt]
\multirow{2}{*}{Co-Training} &	\multicolumn{2}{c|}{Market-1501}	 &	\multicolumn{2}{c|}{DukeMTMC} &	\multicolumn{2}{c|}{CUHK03-Labeled}	&			\multicolumn{2}{c}{CUHK03-Detected}			\\
\cline{2-9}
	&	mAP 	&	rank-1 &	mAP 	&	rank-1  &	mAP 	&	rank-1	&	mAP 	&	rank-1	\\
\hline\hline
No	&	86.4	&	94.4 &	78.0	&	89.0 &	75.1	&	77.2	& 71.8 & 74.2	\\
\hline
Yes &	\textbf{88.2} &	\textbf{95.7} & \textbf{78.9} &	\textbf{89.8} &	\textbf{78.0}	&	\textbf{80.4}	& \textbf{74.8}	& \textbf{77.2}	\\
\bottomrule[1.5pt]
\end{tabular}
\end{center}
\label{tb:co-train}
\end{table*}

With the use of co-training, SDB-2 performs clearly better as shown in Table \ref{tb:co-train} for all the three datasets.   For example, the use of co-training surpass its counterpart by about 3.0\% in mAP for CUHK03.  The motivation behind the use of co-training is that the use of SDB data augmentation for the local branch may hamper the learning of share-net. This suggests that the global branch and the local branch reinforce each other, both contributing to the final performance. It should be pointed out that SDB-2 performs still better than BDB even without use of co-training.

\begin{table}
\caption{Rate of changes for SDB-2 over DukeMTMC-reID}
\begin{center}
\label{my-label}
\begin{tabular}{l||cccc}
\toprule[1.5pt]
$Q$     & mAP & rank-1 & rank-5  & rank-10 \\\hline
1       & 78.8 & 89.5 & 95.1 & 96.5 \\
5       & 78.9 & 89.8 & 95.2 & 96.5 \\
10      & 78.5 & 89.5 & 95.0 & 96.3 \\
\bottomrule[1.5pt]
\end{tabular}
\end{center}
\label{tb:rateofchanges}
\end{table}
\begin{table*}[ht]
\caption{Benefit of Attention Modules}
\begin{center}
\begin{tabular}{l|c@{\hskip 5pt}c@{\hskip 5pt}|c@{\hskip 5pt}c@{\hskip 5pt}|c@{\hskip 5pt}c@{\hskip 5pt}|c@{\hskip 5pt}c@{\hskip 5pt}}
\toprule[1.5pt]
\multirow{2}{*}{Attention Modules} &	\multicolumn{2}{c|}{Market-1501}	 &	\multicolumn{2}{c|}{DukeMTMC} &	\multicolumn{2}{c|}{CUHK03-Labeled}	&			\multicolumn{2}{c}{CUHK03-Detected}			\\
\cline{2-9}
	&	mAP 	&	rank-1 &	mAP 	&	rank-1  &	mAP 	&	rank-1	&	mAP 	&	rank-1	\\
\hline\hline
No	&	87.0	&	95.1 &	77.5  & 88.8 &	77.3 & 79.3	& 73.3 & 75.8	\\
\hline
Yes &	\textbf{88.2} &	\textbf{95.7} & \textbf{78.9} &	\textbf{89.8} &	\textbf{78.0}	&	\textbf{80.4}	& \textbf{74.8}	& \textbf{77.2}	\\
\bottomrule[1.5pt]
\end{tabular}
\end{center}
\label{tb:attention}
\end{table*}
\begin{table}
\caption{Influence of Dropping-Ratio for SDB-2 over DukeMTMC-reID}
\begin{center}
\label{my-label}
\begin{tabular}{l||cccccc}
\toprule[1.5pt]
$r_h$       & 0.1 & 0.2 & 0.3  & 0.4 & 0.5 & 0.6 \\\hline
mAP       & 77.2 & 77.9 & 78.9 & 78.3 & 77.4 & 76.1 \\
rank-1    & 88.8 & 89.6 & 89.8 & 89.4 & 88.6 & 88.4  \\
\bottomrule[1.5pt]
\end{tabular}
\end{center}
\label{tb:erasedh}
\end{table}

\subsubsection{Rate of Changes $Q$ for Slow-DropBlock}
Compared to BDB, the proposed SDB has the freedom to adjust the rate of changes $Q$. The results are shown in Table \ref{tb:rateofchanges} for SDB-2 over DukeMTMC-reID. The use of $Q=5$ can achieve the improvement of 0.3\% on Rank-1 accuracy. However, the use of $Q=10$ does not have any further improvement. This may be due to the fact that the total number of training epochs is fixed, the whole image cannot be fully traversed by randomly-generated dropping patterns when $Q$ increases. For SDB-4, we also observe from experiments that the use of $Q=5$ can achieve consistent performance improvement over that of $Q=1$.

\subsubsection{Benefit of Attention Modules}
With the insertion of attention modules, SDB-2 performs clearly better as shown in Table \ref{tb:attention} for all the three datasets. There is about 1\% improvement on mAP for all the datasets.

\subsubsection{Influence of Dropping-Ratio for SDB}

Table \ref{tb:erasedh} studies the impact of dropping-ratio on the performance of the SDB-2 network. Since the erased width ratio $r_w$ is fixed to 1.0, the dropping-ratio of SDB, defined as the ratio between the dropping area and the area of whole image is fully determined by the erased height ratio $r_h$. We can see that the best performance is achieved when $r_h=0.3$ for DukeMTMC-reID.

\subsubsection{On the Use of Dimension-Reduction Layer}
\begin{table}
\caption{On the use of dimension-reduction (DR) layer for CUHK03-Label}
\begin{center}
\label{my-label}
\begin{tabular}{l||cccc}
\toprule[1.5pt]
Use of DR Layer    & mAP & rank-1 & rank-5  & rank-10 \\\hline
No     & 67.5 & 70.4 & 86.2 & 91.1 \\
Yes    & 78.0 & 80.4 & 92.1 & 95.1 \\
\bottomrule[1.5pt]
\end{tabular}
\end{center}
\label{tb:DRL}
\end{table}

We also perform experiments to investigate the role of dimension-reduction layer. For CUHK03-Label, the results are shown in Table \ref{tb:DRL}. Clearly, A large performance gap is observed and the rank-1 accuracy is 80.4\% for the use of dimension-reduction layer, which surpass its counterpart by 10.0\%. This phenomena, however, is only observed in CUHK03, which is not the case for both Market-1501 and DukeMTMC-reID. Therefore, it deserves further exploitation.

\section{Conclusion}
In this paper, we propose a novel diversity-achieving Slow-DropBlock multi-branch network for person Re-ID. As an alternative structure for the recently proposed Batch DropBlock network, we push the dropping operation to the input images, which, however, has the difficulty for training convergence.  Then, we propose a novel double-batch-split co-training approach with guaranteed training convergence. Experiments show that the proposed diverse SDB network, along with co-training approach, can achieve state-of-the-art performance on several popular person Re-ID datasets, including Market-1501, DukeMTMC-reID and CUHK03.


\end{document}